\def\year{2020}\relax
\documentclass[letterpaper]{article} 
\usepackage{aaai20}  
\usepackage{times}  
\usepackage{helvet} 
\usepackage{courier}  
\usepackage[hyphens]{url}  
\usepackage{graphicx} 
\usepackage{graphicx}  
\usepackage{amsmath,amssymb,amsfonts}
\usepackage{algorithmic}
\usepackage[linesnumbered,ruled]{algorithm2e}
\usepackage{subfigure}
\usepackage[switch]{lineno}
\usepackage{xcolor}
\newcommand\T{\rule{0pt}{2.5ex}}       
\newcommand\B{\rule[-1.2ex]{0pt}{0pt}}

\title{LiDAM: Semi-Supervised Learning with \\Localized Domain Adaptation and Iterative Matching}

\author{\Large Qun Liu, Matthew Shreve, and Raja Bala \\ \\ Palo Alto Research Center \\ Palo Alto, California}

\begin{document}
\maketitle

\begin{abstract}

Although data is abundant, data labeling is expensive. Semi-supervised learning methods combine a few labeled samples with a large corpus of unlabeled data to effectively train models. This paper introduces our proposed method LiDAM, a semi-supervised learning approach rooted in both domain adaptation and self-paced learning. LiDAM first performs localized domain shifts to extract better domain-invariant features for the model that results in more accurate clusters and pseudo-labels. These pseudo-labels are then aligned with real class labels in a self-paced fashion using a novel iterative matching technique that is based on majority consistency over high-confidence predictions. Simultaneously, a final classifier is trained to predict ground-truth labels until convergence. LiDAM achieves state-of-the-art performance on the CIFAR-100 dataset, outperforming FixMatch (73.50\% vs. 71.82\%) when using 2500 labels.

\end{abstract}

\section{Introduction}
Although the amount of data being generated is increasing every year \cite{8776614}, most of this data is unlabeled.  Acquiring labels to sufficiently train fully supervised models is often prohibitively expensive and time-consuming. Given that data-driven deep networks require millions of labeled samples to train, there is a growing interest in semi-supervised methods that learn from a combination of a few labeled samples and a large quantity of unlabeled data \cite{berthelot2019mixmatch,berthelot2019remixmatch,sohn2020fixmatch}. 

We focus on the task of deep image classification. A fundamental component of semi-supervised learning is the detection of clusters that form in the feature projections of unlabeled data. These clusters are assigned pseudo-labels that, when combined with a handful of real labels, can be used to train and update the classifier. The ultimate goal is to group samples together by propagating real labels to nearby samples with a high degree of confidence. In fact, it was recently shown that iteratively applying a combination of clustering and supervised learning (using only pseudo-labels) on a randomly initialized model can achieve remarkable accuracy for image classification \cite{caron2018deep}. In our work, we adopt a similar approach but push performance further by borrowing concepts from transfer learning, and domain adaptation. Additionally our method takes as input initial feature representations from a pretrained or a self-supervised model that requires no human-annotated labels (e.g., MoCo \cite{he2019moco} or SimCLR \cite{chen2020big}). 

\begin{figure}[t!]
\centering
\includegraphics[width=0.48\textwidth]{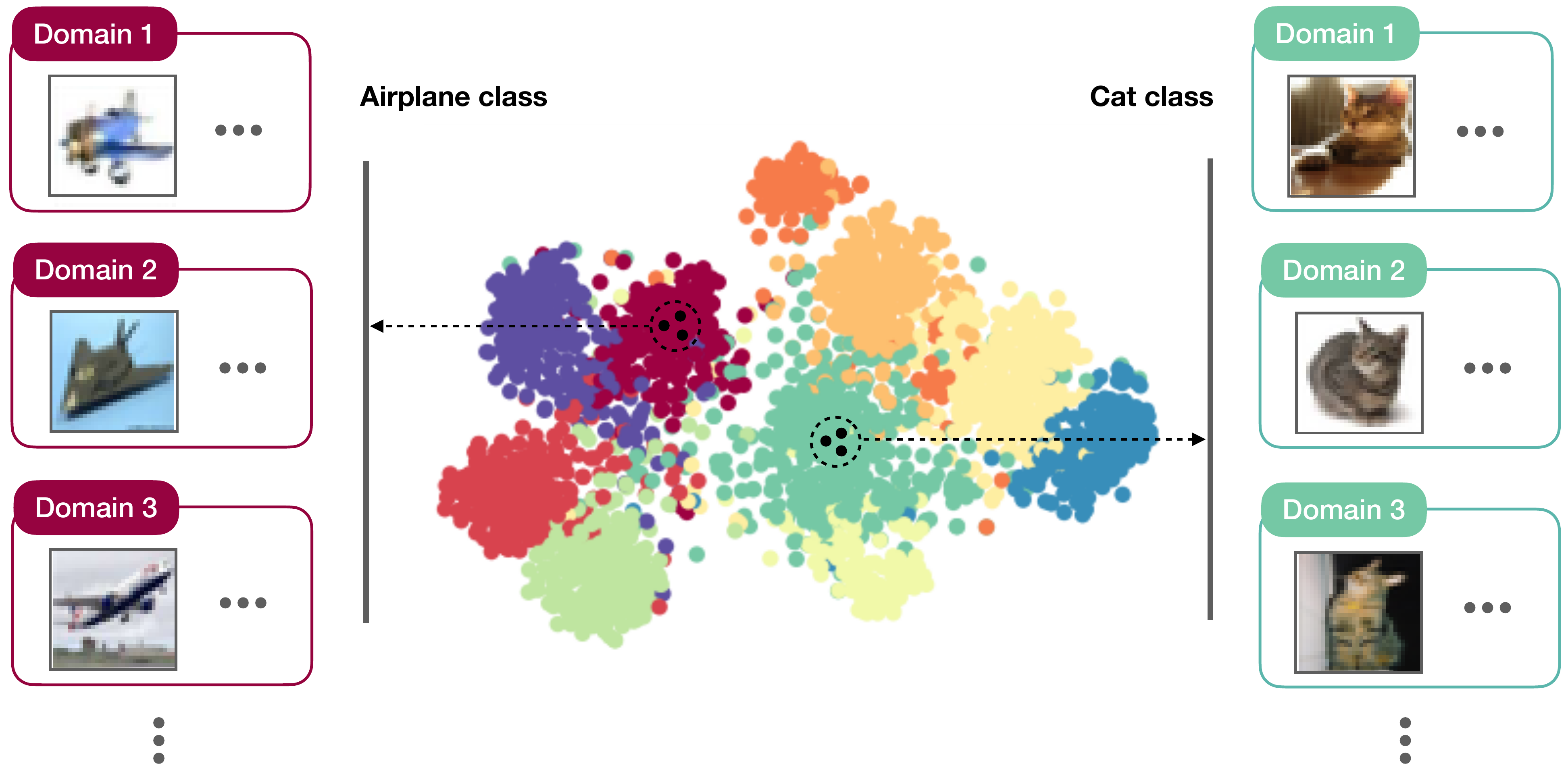}
\caption{Domains of classes. In this figure, we use t-SNE to plot the features extracted from a pretrained imagenet model on the CIFAR-10 training dataset. For each class (represented by a unique color), we select the top k samples (dotted circle). Within this circle, we observe multiple domains for each class. By performing random domain shifts within each of these subregions, we generate domain-invariant feature representations that improve pseudo-labeling and downstream training for classification.
}
\label{fig:lidam_intro}
\end{figure}

Our work is in large part motivated by the observation that many large datasets, especially those collected from online scraping algorithms (e.g., CIFAR, ImageNet, etc.), exhibit high intra-class variance due to diversity in the domains from which the images arise. For example, images within a given class may be sourced from official product advertisements, low quality mobile captures for social networking, images captured from surveillance feeds at oblique poses, etc. In addition intra-class shifts can also arise from regional trends (e.g., images of different dog breeds popular within different populations, but all labeled \textit{dog}). We relieve the impact from intra-class domain variance using localized domain adaptation for learning better essential features of classes.

\begin{figure*}[t!]
\centering
\includegraphics[width=1\textwidth]{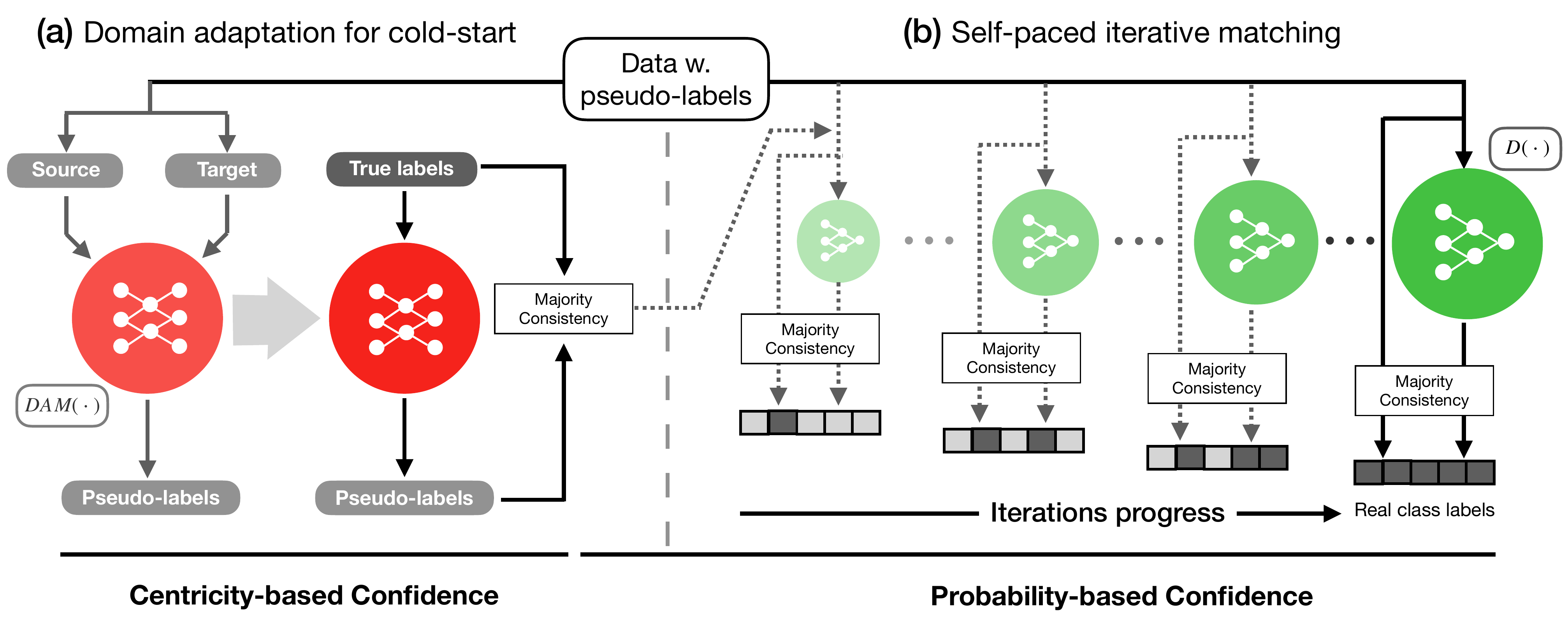}
\caption{Overview of proposed two-stage method. In stage (a), feature points and pseudo-labels from $K$-means are used to train the Domain Adaptation Model (DAM, red circle). DAM generates an initial prediction of class labels, which are fed to stage (b) where labels and the classifier (green circle) are iteratively updated until convergence. Iterations are shown unrolled for illustration (figure best viewed in color).
}
\label{fig:bf}
\end{figure*} 

We propose a two-stage approach as shown in Figure ~\ref{fig:bf}. In the first stage, we begin by generating feature representations from the unlabeled images using a pretrained or self-supervised model. We then apply $K$-means clustering and assign a pseudo-label to each sample based on the cluster ID. For each cluster, a \textit{centricity-based} confidence measure is used to extract two data subsets corresponding to a source and target domain, denoted $S_s$ and $S_t$. Next, a generative domain adaptation model (DAM) is trained to learn domain-invariant feature representations and to predict pseudo-labels in a space with compact clusters. This is followed by an initial alignment between true and pseudo-labels, achieved by feeding labeled samples to DAM, and assigning true labels to all unlabeled data via a \textit{majority consistency} rule. In the second stage, we perform a proposed \textit{self-paced iterative matching} technique to train the final classifier to predict true class labels for all unlabeled samples.

Our novel contributions are twofold: i) the use of deep domain adaptation on artificially induced domain shifts to generate high-quality initial data representations and labels; and ii) introduction of a novel iterative scheme to improve at once the data labels and the final classifier.

Our experimental results demonstrate that our method can align pseudo-labels to ground-truth labels with a high degree of accuracy. Overall, our method achieves competitive performance on both the CIFAR-10 and CIFAR-100 datasets under the same semi-supervised setting. On the CIFAR-100 dataset, we outperform the state-of-the-art (73.50\% vs. 72.88\%) when only using 5\% of the available labels, while outperforming the majority of methods when a larger number of labels is used. To the best of our knowledge, LiDAM is the first approach that combines unsupervised clustering with localized domain adaptation and an iterative matching mechanism to generate accurate pseudo-labels.

\section{Related work}

\noindent\textbf{Domain adaptation and transfer learning}
Recently, adversarial learning has been used to improve the marginal and conditional distributions of cross-domain sample populations for the purpose of transfer learning \cite{yu2019transfer,zhao2018adversarial}. Such approaches jointly train a feature generator ($G_f$) and discriminator ($G_d$) so that the final distributions between the source and target domains are as close as possible, resulting in a domain invariant set of features. \cite{Lee_2019_ICCV} improves on this idea by proposing an additional filter that ignores portions of visual data that do not transfer well through the use of a \textit{residual transferability aware bottleneck}. Alternatively, negative and positive transfer measures can determine how effective features learned for some previous task transfer to a new target task. In the work by \cite{cao2018partial}, these measures are explored on each class separately to determine if a class should be learned in the source domain and if so, how it should be weighted in a transfer process. In contrast to these approaches which seek to generally globally adapt inter-class features across two domains, we locally adapt features learned from the unaltered source domain by locally adapting features learned for each class to suppress intra-class domain variances.

\noindent\textbf{Semi-supervised learning}
The sample-efficient method proposed in \cite{lee2013pseudo} assigns pseudo-labels to unlabeled data for classes that are predicted with high probability. These labels are then aligned with true labels (using a majority rule) to train a proposed network in supervised fashion. Recent work has dramatically achieved better performance for semi-supervised learning including the work by \cite{berthelot2019mixmatch} which introduces the MixMatch algorithm, which proposes the notion of using low-entropy labels for unlabeled data. Their approach also mixes augmented data with labeled data to predict the classes of unlabeled data. Their approach was later improved in ReMixMatch \cite{berthelot2019remixmatch} by aligning the distributions ofaugmentation anchoring techniques, to align marginal distributions between predictions of unlabeled data and true labels and to make the output of strongly augmented input close to that of the weakly augmented one. Same authors \cite{sohn2020fixmatch} proposed FixMatch simply combines consistency regularization and pseudo-labeling to align the predictions of pseudo labels between weakly and strongly augmented unlabeled images. Our approach is different from above. In general, we initiate our model with pseudo labels from unsupervised clustering, then align pseudo labels to true labels using our proposed iteratively matching based on cold start from localized domain adaptation (see more details in Section \ref{iterative}).

\begin{figure}[t]
\centering
\includegraphics[width=0.47\textwidth]{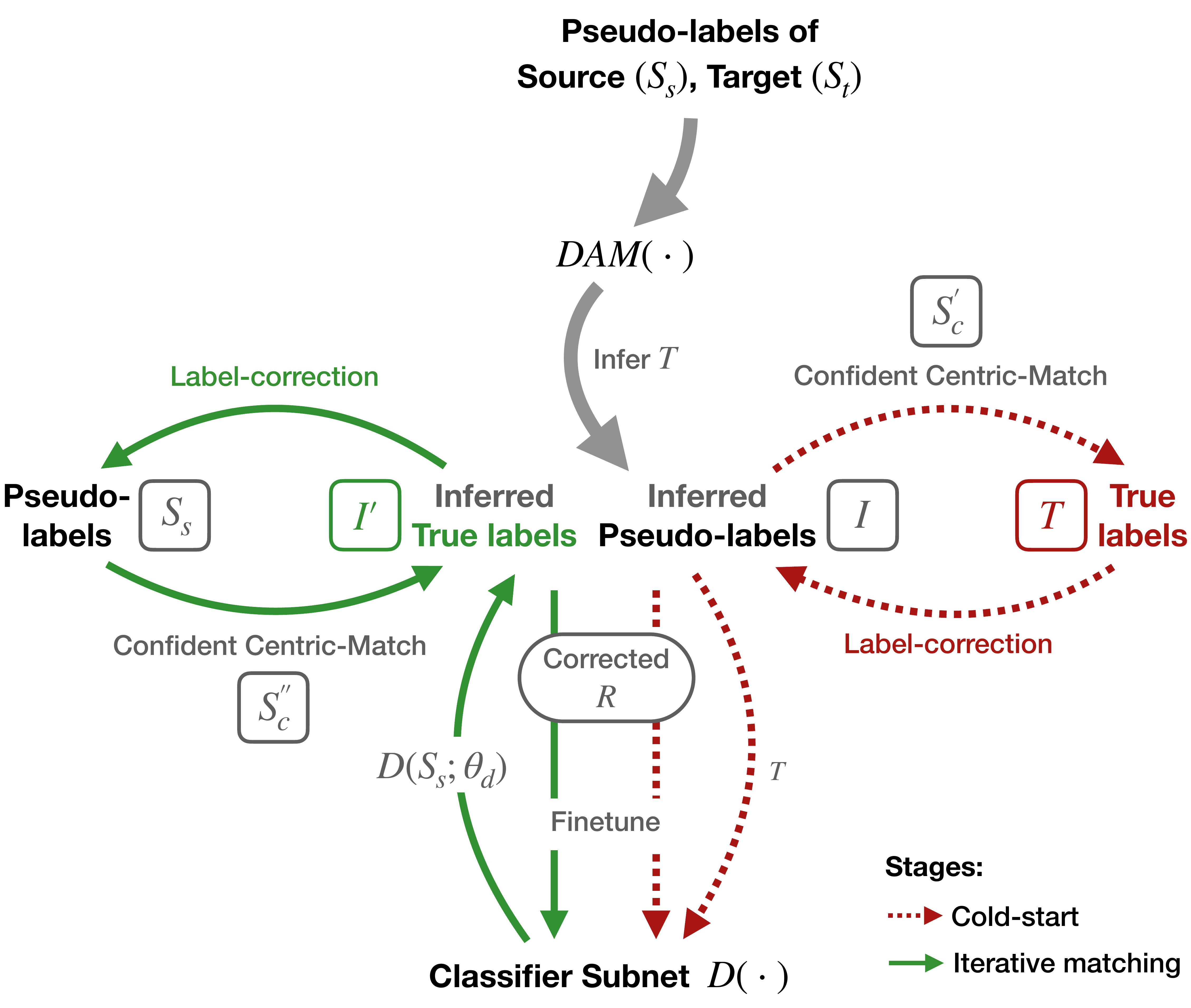}
\caption{Detailed illustration of our proposed approach. A domain adaptation model trained on pseudo-labels is initially used to predict real class labels for an unlabeled dataset $T$, solving the cold start problem (dashed loops in red). These predictions are then used initially train a separate classifier $D$ that is iteratively updated using alternating label mapping mechanisms (solid green loops).}
\label{fig:algo}
\end{figure}

\section{Method} \label{method}
In this section, we describe in detail our proposed LiDAM approach. We acquired pseudo-labels for unlabeled datasets using unsupervised clustering approaches (in Section \ref{uc}). Then, we introduce the domain adaptation model in self-supervised training to solve cold-start problem (in Section \ref{dam}). Finally, iteratively matching mechanism for mapping pseudo-labels to true labels introduces in Section \ref{initialize} and \ref{iterative}. Our proposed method is shown in the Figure \ref{fig:algo}.

\subsection{Unsupervised Clustering} \label{uc}
We perform \textit{K}-means clustering on features extracted from each image and assign an initial pseudo-label prediction based on cluster ID. Rather than cluster on features from a randomly initialized model as performed in \cite{caron2018deep}, we cluster on the feature space generated from either a model pretrained on a different dataset, or one trained using a self-supervised learning pretext task \cite{he2019moco}. Next we randomly split the high-confidence central nearest points into two subsets that we treat as source and target domains. Since the compact groups of samples that tightly surround each cluster's mean location are more likely belong to the same class, as demonstrated in \cite{garg2017supervising}. In addition, as we observe that even though these points have a high likelihood of belonging to the same class, they also belong to different domains (recall that images of a car that could be taken from cameras, phones, product advertisements, or other device resources). Here, we do not have a distance-based heuristic for deciding source and target domains. Instead, within the center-nearest points, we select the majority of points as a dominant domain/source domain since we believe these data points have a high probability of capturing most of the generalizable features. We reduce the impact from each intra-class domains shift (whatever an image of a car is cartoon or real) by forcing the model sufficiently learn cross-domain features over all data points. These improved features are then subsequently used to more accurately assign labels to low-confidence samples outside of this fixed radius in the future iterative matching stage.

\subsection{Domain Adaptation Model} \label{dam}
Several recent domain adaptation techniques attempt to learn domain-invariant features from data arising from multiple domains. We adopt the adversarial domain adaption strategy described in \cite{ganin2015unsupervised} comprising 3 subnets. First, a subnet $M(\cdot)$ with parameters $\theta_m$ learns domain-invariant feature representations. These are then fed to both a domain discriminator $G(\cdot)$ with parameters $\theta_g$ that distinguishes between source and target domain; and a classifier $C(\cdot)$ with parameters $\theta_c$ that predicts class labels. Training proceeds as a two-player \textit{minmax} game that simultaneously maximises the loss $\mathcal{L}_{g}$ of $G(\cdot)$ while minimizing the loss $\mathcal{L}_{c}$ of  $C(\cdot)$. The overall network loss is given by:

\begin{equation}
\begin{aligned} 
\mathcal{L}_{ori}(\theta_m, \theta_c, \theta_g) 
&=\frac{1}{N_s} \sum_{i=1}^{N_s} \mathcal{L}_{c} (y_i, C(M(X_i)))\\
&-\frac{\lambda}{N} \sum_{j=1}^{N} \mathcal{L}_{g} (l_j, G(M(X_j)))
\end{aligned} 
\end{equation}

\noindent where $X_i$ and $y_i$ are input images and pseudo-labels respectively, $N_s$ and $N_t$ are the number of samples in the source and target domains respectively, and $N$ is the total dataset size. Hyperparameter $\lambda$ balances the importance of classifier loss $\mathcal{L}_{c}$  vs. domain discriminator loss $\mathcal{L}_{g}$. The binary variable $l_j \in \{0, 1\}$ denotes whether a sample is from the source  or target domain. Network parameters are learned through the \textit{min-max} optimization shown below,

\begin{equation}
\begin{aligned} 
(\hat{\theta}_m, \hat{\theta}_c) 
&=\operatorname*{arg\,min}_{\theta_m, \theta_c} \mathcal{L}_{ori}(\theta_m, \theta_c, \theta_g), \\
(\hat{\theta}_g) 
&=\operatorname*{arg\,min}_{\theta_g} \mathcal{L}_{ori}(\theta_m, \theta_c, \theta_g). 
\end{aligned} 
\end{equation}

Since the overall number of domains present in each class is unknown and therefore we do not know how many will fall into the source and target domain sets, we adopt a partial domain adaptation technique. In general, aligning the source label space $L_s$ to the target label space $L_t$ can result in classification performance decay due to effects from the outlier label space (subspace of source labels space that are not represented in the target domain). We define $L_s\backslash L_t$ as a target label space that is a subspace of source label space, where $|L_t|\ll|L_s \backslash L_t|$ since transferring from a large source set to a small target set is a normal scenario (this is reflected in our choice of source and domain set sizes in Section 4). We implement the partial adversarial domain adaptation approach defined in \cite{cao2018partials} by assigning weights on source domain classes to reduce the interference from other classes when learning transferable features. The weight $\mathbf{v}$ is a $|L_s|$ dimensional vector to augment weights on $L_s\cap L_t$ which forces the weights on $L_s\backslash L_t$ to be very small. Specifically, $\mathbf{v}$ quantifies contributions from each class by averaging the label predictions over target data,

\begin{equation}
\begin{aligned} 
\mathbf{v} = \frac{1}{N_t} \sum_{i=1}^{N_t} \hat{\mathbf{y}}_i,
\end{aligned} 
\end{equation}

Note that the weight $\mathbf{v}$ satisfies $\sum_{i=1}^{|L_s|} v_i=1$. By normalizing the weight $\mathbf{v}$ by its maximum value, $\mathbf{v}/\max(\mathbf{v})$ learns to assign small valued weights on $L_s\backslash L_t$. We apply the weight $\mathbf{v}$ to the source classifier $C(\cdot)$ and the discriminator $G(\cdot)$ over source domain data to equations $(1)$ and $(2)$ and derive the partial domain adaptation,

\begin{equation}
\begin{aligned} 
\mathcal{L}_{par}(\theta_m, \theta_c, \theta_g) 
&=\frac{1}{N_s} \sum_{i=1}^{N_s} v_{y_i} \mathcal{L}_{c} (y_i, C(M(X_i)))\\
&-\frac{\lambda}{N_s} \sum_{i=1}^{N_s} v_{y_i} \mathcal{L}_{g} (l_i, G(M(X_i))) \\
&-\frac{\lambda}{N_t} \sum_{j=1}^{N_t} \mathcal{L}_{g} (l_j, G(M(X_j)))
\end{aligned} 
\end{equation}

\noindent where the hyper-parameter $\lambda$ is same as defined in equation $(1)$ and $v_{y_i}$ is the weight corresponding to the ground-truth label $y_i$ with source data $X_i$. The parameters are then learned through \textit{minmax} optimization thus shown below,

\begin{equation}
\begin{aligned} 
(\hat{\theta}_m, \hat{\theta}_c) 
&=\operatorname*{arg\,min}_{\theta_m, \theta_c} \mathcal{L}_{par}(\theta_m, \theta_c, \theta_g), \\
(\hat{\theta}_g) 
&=\operatorname*{arg\,min}_{\theta_g} \mathcal{L}_{par}(\theta_m, \theta_c, \theta_g). 
\end{aligned} 
\end{equation}

\begin{algorithm}[t!]
\caption{Iterative Matching} \label{alg1}
Pretrain($DAM$)\\
$DAM(T;\theta_c) \rightarrow I \hspace{35pt}\blacktriangleright$ inferred pseudo-labels\\
$I \rightarrow S_c^{'} \hspace{40pt}\blacktriangleright$ select confident pseudo-label set\\
Initialize($Loop$) \\
\While{$Acc_L>=Acc_{best}$}
{   
    \uIf {$Model_L$ not exist}
    {   
        \For{number of classes}
        {
            $l_{new} = Map(\mathbf{I}^{l_p}\rightarrow \mathbf{T}^{l_t}) \blacktriangleright$ match to true labels\\
            $Update(R) \gets l_{new}$ \\
        }
    }
    
    \Else
    {   
        $Acc_{best} = Acc_L$ \\
        $Model_{best} = Model_L$ \\
        $D(S_s;\theta_d) \rightarrow I' \hspace{30pt}\blacktriangleright$ inferred true labels\\
        $I' \rightarrow S_c^{''} \hspace{10pt}\blacktriangleright$ confident inferred true label set\\
        \For{number of classes}
        {
            $l_{new}^{'} = Map(\mathbf{S}_s^{l_p^{'}}\rightarrow \mathbf{I'}^{l_t^{'}}) \blacktriangleright$ match to inferred true labels\\
            $Update(R) \gets l_{new}^{'}$ \\
        }
    }
    Tune classifier $D(R;\theta_d)$ and finetune with $T$ \\
    Save$(Model_L, Acc_L) \hspace{13pt}\blacktriangleright$ save model, accuracy
}
\end{algorithm}

\subsection{Label initialization} \label{initialize}
Given an initially pseudo-labeled training set $R$, we identify a subset of samples $S_c$ for which we have a high degree of confidence in the pseudo-label prediction. This is achieved by selecting the $k$ neighbors nearest to each cluster centroid based on Euclidean distance. $S_c$ is randomly divided into a source domain set $S_s$ and a target domain set $S_t$. We report the performance of LiDAM using multiple choices of $k$ that are selected based on balancing psuedo-label and ground-truth label prediction (this is explained in more detail in Section 4 and Figure~\ref{fig:damfig}).

To start the iterative matching process and address the cold-start problem of predicting ground-truth labels for unlabeled samples, we adopt the following  \textit{majority consistency} approach. We run the small set $T$ of labeled samples through $DAM(\cdot)$ to get predicted pseudo-labels. Note that samples from a given true class can map to multiple pseudo-labels. We count the occurrences of pseudo-label predictions within a given true class, and assign to that class the pseudo-label with the majority count. Now every labeled sample has both a true and pseudo label. Next, the loop is closed by reversing this mapping and aligning each pseudo-label with a ground-truth label for all unlabeled samples, as shown in the right half of Figure~\ref{fig:algo}. To this end, we define a function $Map$ that matches pseudo-class label $l_p$ to a true class label $l_t \in [0, Z]$  and returns the true label $l_t$ as correct label $l_{new}$, 

\begin{equation}
l_{new} = Map(\mathbf{I}^{l_p}\rightarrow \mathbf{T}^{l_t}).
\label{eq:map}
\end{equation}

To do this, $Map$ first counts the frequency of the ground-truth labels associated with each group of samples that have the same high-confidence pseudo-label prediction from $DAM(\cdot)$. Initially, all high confidence pseudo-label predictions are stored in $S_c^{'}=\{x_i^j | P_r(I_i)>=1-\alpha, i\in 1,2,...,|T|, j\in 1,2,...,Z\}$ where $\alpha$ is a hyper-parameter ($\alpha=1\mathrm{e}{-5}$ in our experiments). Then, all samples $S_c^{'}$ with the same psuedo-label prediction $l_p$ are collected in the set $\mathbf{x}^{l_p}\in S_c^{'}$. After determining the most frequently occurring ground-truth label $l_t$, we propagate the ground-truth label to all members of $\mathbf{x}^{l_p}$ by replacing the pseudo-label with the true class label $l_t$. After each pseudo-class label has been matched and updated in $R$, then we use this newly labeled dataset to train a new classifier $D(\cdot)$ (ResNet-50) which is subsequently finetuned with $T$ for downstream classification.

\subsection{Iterative matching} \label{iterative}
After the cold-start initialization using $T$, $D(\cdot)$ is able to start inferring true class labels $I'$ for $S_s$ (the initial high centricity-based confidence source set). During each iteration (denoted as a solid green line in Figure~\ref{fig:algo}), the assigned pseudo-label $S_s$ is updated based on high confidence predictions by $D(\cdot)$. To achieve this, the mapping rule as described in is in the reverse direction as described in Eq.~\ref{eq:map} since $D(\cdot)$ predicts true labels as opposed to the pseudo-labels predicted by $DAM(\cdot)$. Thus, $Map$ follows the same process but maps each pseudo-label $l_p^{'}$ of $S_s$ to a true class label $l_t^{'}$ of $I'$ and returns the true class label as the correct label $l_{new}^{'}$,

\begin{equation}
l_{new}^{'} = Map(\mathbf{S}_s^{l_p^{'}}\rightarrow \mathbf{I'}^{l_t^{'}}).
\end{equation}

\noindent where $l_t^{'}$ is the most frequent inferred true label on $\mathbf{x}^{l_p^{'}}$ with associated pseudo-label $l_p^{'}$, and $\mathbf{x}^{l_p^{'}}\in S_c^{''}$ and $S_c^{''}$ is an inferred true label set that contains samples with high probability detections from $D(\cdot)$. That is, $S_c^{''}=\{x_i^j | P_r(I_i^{'})>=1-\alpha', i\in 0,1,2,...,|S_s|, j\in 0,1,2,...,Z\}$ where $\alpha'$ is a hyper-parameter that thresholds predicted confidence values ($\alpha'=1\mathrm{e}{-2}$ in our experiments). Note that $S_s$ can potentially affect downstream classification performance if the initial \textit{K}-means clustering performance is poor. Thus, choosing a good size for $S_c$ in critical (we explore different sizes for $S_s$ in our experimental results and in Table~\ref{table:kcs10} and Table~\ref{table:kcs100}). 
Finally, after pseudo-class labels are matched and updated in $R$ with $T$, we again finetune the classifier $D(\cdot)$ and continue iterating. As the number of iterations increase, more and more pseudo-labels will be correctly aligned to true labels (as illustrated in Figure~\ref{alg1}-b). The classifier $D(\cdot)$ is thus progressively improved by optimizing the following problem,

\begin{equation}
\begin{aligned} 
\operatorname*{arg\,min}_{\theta_d} \sum_{i=1}^{|R|} \mathcal{L}_{d} (y_i, D(x_i;\theta_d))
\end{aligned} 
\end{equation}

where $\theta_d$ is the parameters of classifier $D(\cdot)$ with the loss $\mathcal{L}_{d}$ for $x_i\in R$. The iterative matching algorithm is provided in Algorithm~\ref{alg1}. At the end of our approach, the best local model $D(\cdot)$ is used for inferring true labels for all samples until the predicted class labels converge with the ground-truth labels for all samples in $T$. 

\begin{table}[t]
\caption{The effect from the size of confident pseudo label set for the performance of Kmeans clustering on CIFAR-10.}
\vspace{5pt}
\begin{center}
\begin{tabular}{ccc}
    \hline
 Cluster subset $|S_c|$ & Total correct & \textit{K}-means acc. \T\B \\ 
 \hline
 \T 500 & 4762 & 0.9524 \\ 
 1000 & 9326 & 0.9326  \\ 
 1500 & 13732 & 0.9155  \\ 
 \B 2000 & 18089 & 0.9045 \\ 
\hline
\end{tabular}
\label{table:kcs10}
\end{center}
\end{table}

\section{Experiments}
We demonstrate the performance of our method on image classification, making comparisons to several state-of-the-art methods. Our experiments are performed on two benchmark image classification datasets, CIFAR-10 and CIFAR-100\cite{krizhevsky2009cifar} using standard ground-truth partitioning techniques used for measuring semi-supervised learning performance. 

\subsection{Datasets}

\textbf{CIFAR-10} contains a total of 60000 RGB images from ten classes. The image size is 32$\times$32 and each class has 6000 images. 1000 images from each class are randomly selected to form a test set, while the remaining images are used for training.

\noindent\textbf{CIFAR-100} is similar to CIFAR-10 except with 100 classes, each containing 600 RGB images. 500 images from each class comprisd a training dataset of 50000 images and the remainder are used for testing. 
CIFAR-100 is more challenging as it has much more classes and fewer images for training than CIFAR-10.

\begin{figure*}[t!]
\centering
\subfigure[Performance at various source and target sizes ($S_c$) on CIFAR-10 using pseudo-labels]{
    \includegraphics[width=0.31\linewidth, keepaspectratio]{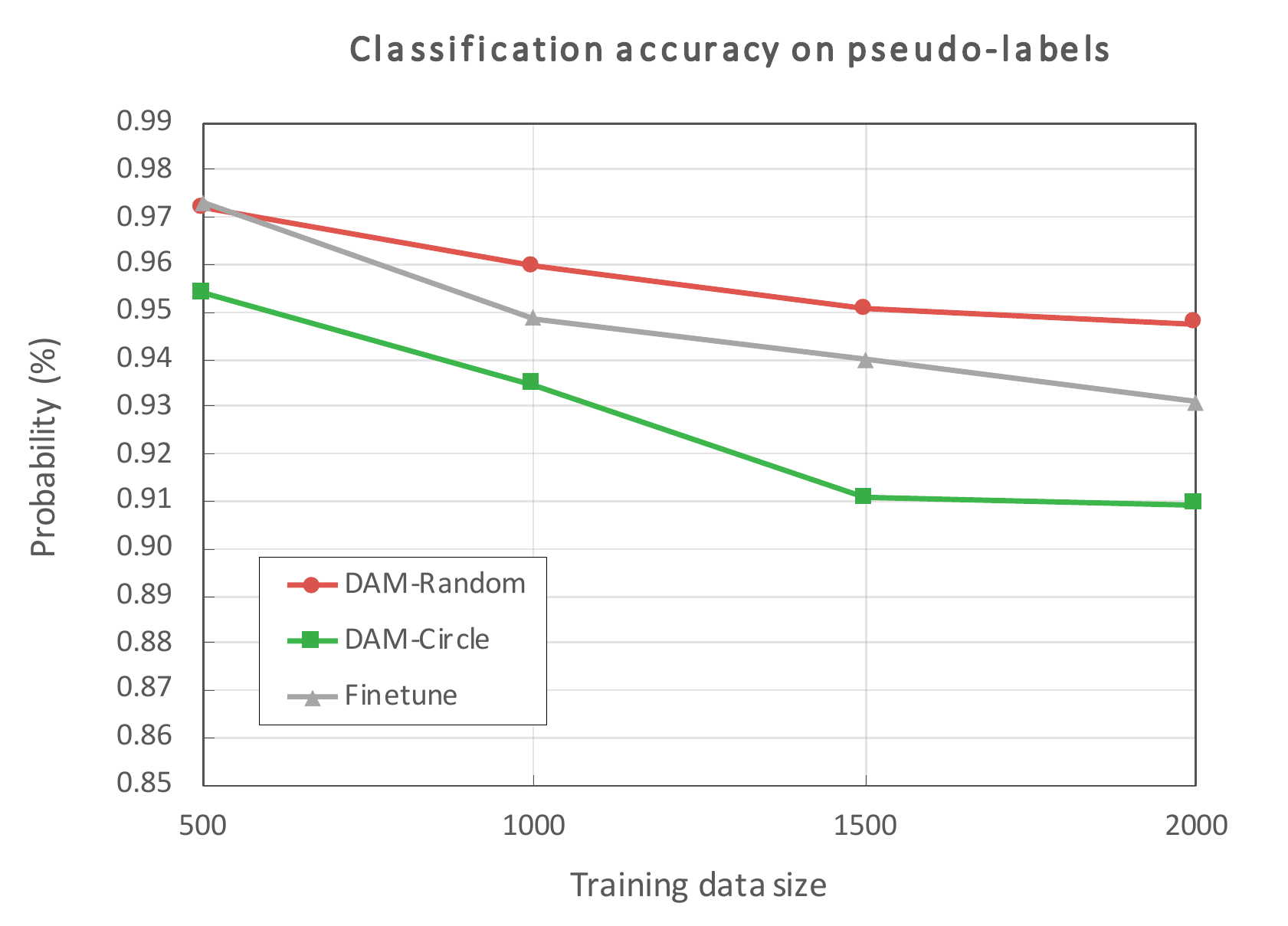}
    \label{fig:a}
}
\hspace{5pt}
\subfigure[performance at various source and target sizes ($S_c$) on CIFAR-10 using ground-truth labels]{
    \includegraphics[width=0.31\linewidth, keepaspectratio]{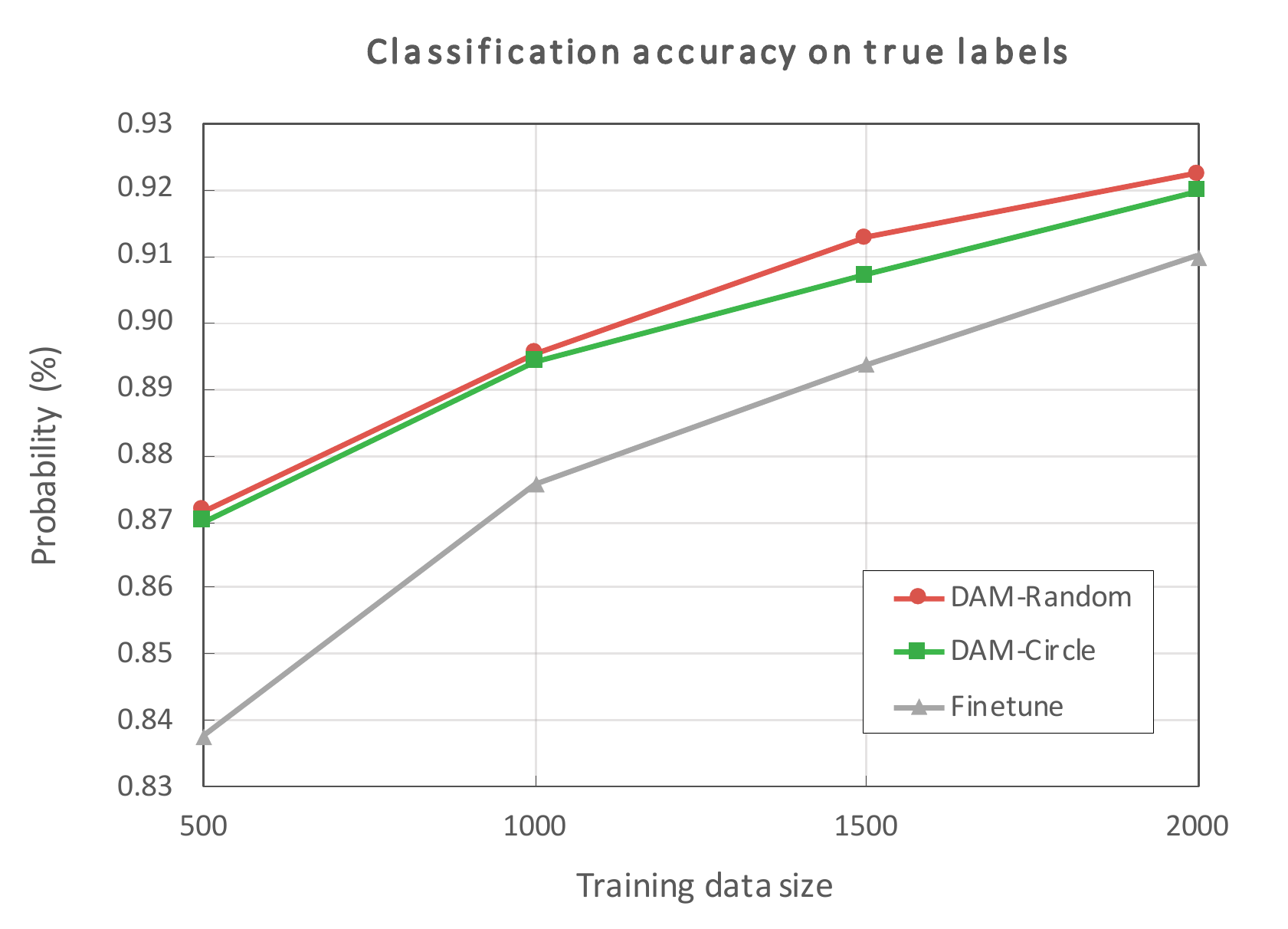}       
    \label{fig:b}
}
\hspace{5pt}
\subfigure[performance at various source and target sizes ($S_c$) on CIFAR-10 and using test data from CIFAR-10]{
    \includegraphics[width=0.31\linewidth, keepaspectratio]{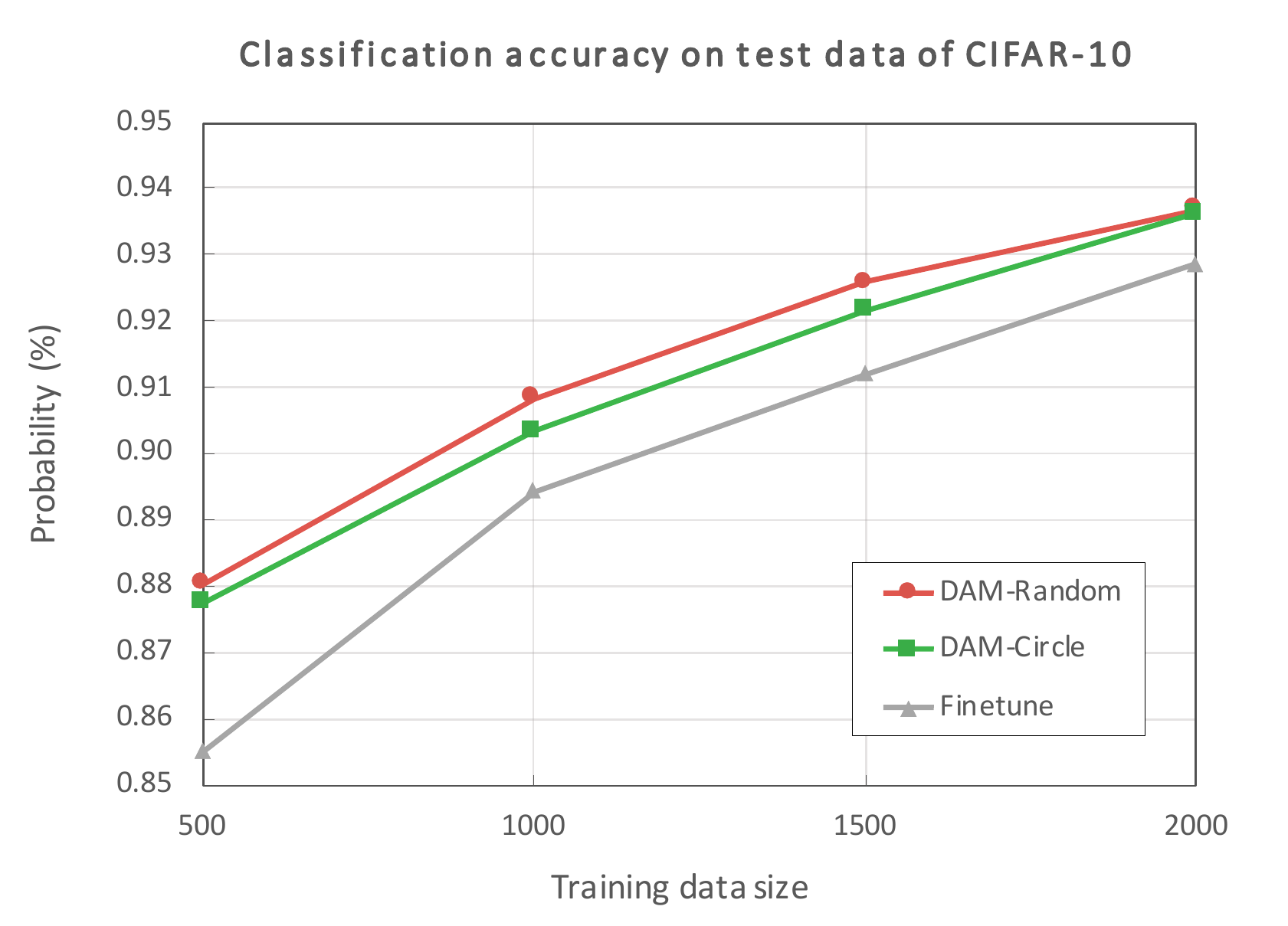}       
    \label{fig:c}
}

\caption[Optional caption for list of figures]{Explorations and comparisons for pretraining DAM. These experiments test both the size of $S_c$ as well as how $S_c$ is partitioned into the source ($S_s$) and target ($S_t$) subsets. }
\label{fig:damfig}
\end{figure*}

\begin{table}[t!]
\caption{The effect of the size of confident pseudo label set for the performance of Kmeans clustering on CIFAR-100.}
\vspace{5pt}
\begin{center}
\begin{tabular}{ccc}
    \hline
Cluster subset $|S_c|$  & Total correct & Kmeans acc. \T\B \\ 
 \hline
 \T 30 & 1775 & 0.5917 \\ 
 60 & 3415 & 0.5692 \\ 
 90 & 5052 & 0.5613 \\ 
 \B 120 & 6553 & 0.5461 \\ 
\hline
\end{tabular}
\label{table:kcs100}
\end{center}
\end{table}

\subsection{Implementation details}

Our approach relies on an initial set of feature representations that are used in two different stages: (1) in the initial feature extraction and clustering stage; (2) in the classifier that is finetuned during the iterative matching stage. We test our approach by using a ResNet-152 pretrained on ImageNet for the initial feature extracting and clustering stage, and ResNet-50 pretrained on ImageNet as the backbone for the classifier in the iterative matchinng stage. 

To collect the source and target domain datasets, we randomly divide $S_c$ using an 80 / 20 split. For selecting the number of ground-truth samples used to train our models, we follow standard practice: for CIFAR-10, the test on ground-truth selections sizes of 250, 1000, and 4000 which comprise images randomly selected from each class in groups of 25, 100, and 400, respectively; for CIFAR-100, ground-truth selections of 2500, 5000, and 10000 are comprised by randomly selecting 25, 50, and 100 images per class, respectively. For our reported image classification tests, we use the standard test data partition from CIFAR-10 and CIFAR-100. Experiments on CIFAR-10 conducted on 3 Tesla V100 GPUs and experiments on CIFAR-100 conducted on 3 GTX 1080 Ti GPUs.

\subsection{Evaluation}

We first evaluate how well we \textit{K}-means clustering performs when choosing the closest \textit{N} samples around each cluster center $S_c$. As can be seen from Table~\ref{table:kcs10} and Table~\ref{table:kcs100}, different choices of \textit{N} lead to different levels of cluster accuracy or purity (i.e., the number of samples with a common ground-truth label that is unique from other clusters). One intuitive but clear takeaway from both of these results is that as the distance increases from the cluster center, the more likely we are to find samples of different classes. As discussed in Section 3, this is important for two reasons. The first reason is that since the subset $S_c$ is used for selecting the intra-class source and target domain sets around each cluster which are used to train $DAM$, any impurity will likely increase inter-class confusion. The second reason is that these samples are expected to be of a singular class when training the final classification model during the iterative matching stage.

\begin{table*}[!t]
    \caption{Comparison of classification accuracy on CIFAR-10 and CIFAR-100 with baseline models such as $\Pi$-model \cite{laine2016temporal}, PseudoLabel \cite{lee2013pseudo}, MixUp \cite{zhang2017mixup}, VAT \cite{miyato2018virtual} MeanTeacher \cite{tarvainen2017mean}, MixMatch \cite{berthelot2019mixmatch}, EnAET \cite{wang2019enaet}, UDA \cite{xie2019unsupervised}, ReMixMatch \cite{berthelot2019remixmatch}, and FixMatch \cite{sohn2020fixmatch}. The highest reported accuracies are used for baseline models. Top accuracies are reported in bold, including those that fall within the reported error margins.}
    \vspace{10pt}
    \begin{minipage}{.5\linewidth}
      \centering
        \begin{tabular}{lcccc}
        \hline
        \multicolumn{4}{c}{CIFAR-10} \T\B \\ \hline
        Method & 250 labels & 1000 labels & 4000 labels \T\B \\ 
        \hline
         \T $\Pi$-model & 0.4903 & 0.6945 &0.8296\\ 
         PseudoLabel & 0.5119 & 0.7082 &0.8390\\ 
         MixUp & 0.5349 & 0.7494 &0.8705\\ 
         VAT & 0.6679 & 0.8172 &0.8926\\ 
         MeanTeacher &0.5739 &0.8668&0.8989\\
         MixMatch &0.8979 & 0.9257&0.9382\\
         ReMixMatch &\textbf{0.9461} & - &0.9541\\
         \B FixMatch (RA) &\textbf{0.9558} & - &\textbf{0.9579} \\
         \hline
         LiDAM & 0.8083 & 0.8904 & 0.9252 \T\B \\
        \hline
        \end{tabular}
    \end{minipage}%
    \begin{minipage}{.5\linewidth}
      \centering
        \begin{tabular}{lcccc}
        \hline
        \multicolumn{4}{c}{CIFAR-100} \T\B \\ \hline
        Method & 2500 labels & 5000 labels & 10000 labels \T\B \\ 
        \hline
         
         \T $\Pi$-model & 0.4323 & - &0.6223 \\ 
         PseudoLabel & 0.4308 & - &0.6398 \\ 
         Mean Teacher & 0.4666 & - &0.6441 \\ 
         MixMatch & 0.6043 & - &0.7202 \\ 
         EnAET & - &0.6817 &0.7328 \\
         UDA &0.6709 & - &0.7575 \\
         ReMixMatch & 0.7288 & - & \textbf{0.7753} \\
         \B FixMatch (RA) &0.7182 & - &\textbf{0.7752} \\
         \hline
         LiDAM & \textbf{0.7350} & \textbf{0.7514} & \textbf{0.7678} \T\B \\
        \hline
        \end{tabular}
    \end{minipage}
    \label{table:result}
\end{table*}

To get an insight of how the choice of $S_c$ and its subdivisions $S_t$, and $S_s$ will effect the training performance of DAM, we conducted a series of experiments as shown in Figure \ref{fig:damfig}. In these experiments, $S_t$ serves as test data except for the experiment shown in Figure \ref{fig:c} where we use the standard CIFAR-10 test data partition. We propose two approaches for subdividing $S_c$ into the source ($S_s$) and target ($S_t$) domains. In the first approach, each sample is randomly assigned to a domain based on a predefined ratio (e.g., 80/20 split). This approach captures the intuition that multiple domains are spread across the each cluster, and so features can be best extracted after performing multiple local domain shifts in random directions. In the second approach, the distance to the cluster center is used to separate the two domains. This is defined by selecting points that fall within or outside an inscribed circle/sphere (i.e., the closest 80\% are assigned to the source domain, the remaining 20\% to target domain). The intuition behind this approach is that sample points nearest to cluster centroids will likely belong to a single domain; thus features are best extracted by shifting outward to domains in the periphery. As a baseline approach, we compare the perfomance of each of these experiments to a ResNet-50 model finetuned on $S_s$.

In Figure~\ref{fig:a}, we observe the performance of the DAM model at predicting psuedo-labels using both of these partitioning approaches on $S_t$. They key takeaway from this plot is that randomized domain shifts (DAM-Random) result in the extraction of more robust features compared to DAM-Circle and Finetune as the size of $S_c$ increases and it's purity decreases (see Table~\ref{table:kcs10}). For comparison in Figure~\ref{fig:b}, we use the same partitions used in Figure~\ref{fig:a}, but instead train and predict on ground-truth labels. Here we can see that the model does not suffer from poor \textit{K}-means psueudo-label assignment and instead leverages the true labels to progressively improve performance as more of it is provided. Finally, in Figure~\ref{fig:c} we run this same experiment but test performance on the CIFAR-10 test partition. Overall, we conclude from these experiments that the DAM-Random approach performs better than the DAM-Circle approach, $DAM$ is more accurate at predicting pseudo-labels with smaller sizes (e.g., 500) of $S_c$, and lastly $DAM$ performs better than a finetuned ResNet-50 for learning essential features across multi-source domains. Therefore, we select this configuration and combine it with our iterative matching algorithm test our full LiDAM framework.

\textbf{CIFAR-10 Results} We compare our results with other methods using varying amounts of ground-truth training data. We follow standard practice by randomly sampling 25, 100, and 400 labels per class. A full comparison of results is provided in Table \ref{table:result}. We provide the highest accuracies reported for each method by the original authors. Top accuracies are indicated in bold, including those that fall within the reported error margins. Our approach is competitive but is outperformed in all labeling cases for CIFAR-10.

For 250 labels provided from CIFAR-10, FixMatch \cite{sohn2020fixmatch} achieved the overall best accuracy 0.9558 which is slightly better than ReMixMatch \cite{berthelot2019remixmatch} and than our method. Our method performs better on 1000 labels and 4000 labels, but still outperforms most of methods on 250 labels such as MeanTeacher, VAT, etc. For 1000 labels, MixMatch achieved best accuracy 0.9257, outperforms our method by 3.53\%. Our method obtains accuracy of 0.8904 which has 2.36\% improved than the third best method. For 4000 labels, our method obtained 0.9252 while the best method FixMatch obtained 0.9579. As we can see from the table, even only 1000 labels provided, our method is also competitive to the most of other methods with 4000 labels.

\textbf{CIFAR-100 Results} We also compare our results on the CIFAR-100. Results are also provided in Table \ref{table:result}. Our proposed method outperforms all other methods on 2500 labels and 5000 labels.  On 2500 labels, we observe a 0.6\% improvement compared to ReMixMatch and 1.68\% improvement compared to FixMatch. Similar to the results on CIFAR-10, we outperform most other methods when using the minimal amount of data (2500 labels) even when other approaches use the all available data (10000 labels). Our method achieves competitive accuracies on 10000 labels but ultimately falls short of ReMixMatch (0.7641).

Overall, our method achieves state-of-the-art accuracies for CIFAR-100, but performs worse on CIFAR-10. Our main hypothesis for this discrepancy is that for the CIFAR-100 case, the number of samples in each class are small but the domain variance is large. This is likely an optimal scenario for LiDAM which primarily address such domain variance. On the other hand, CIFAR-10 contains fewer classes but a large number of samples per class. Therefore, there is an increased chance that the clusters identified by \textit{K}-means will have accurately identified all samples across all domains, leading to less domain variance in the regions selected around each cluster centroid. Another point worth noting is that that when using a large number of labels such as 10000 labels from CIFAR-100, as shown in Table \ref{table:result}), our method approaches the top accuracy but falls slightly short. One explanation for this small gap is our choice to use ResNet-50 as our backbone, opposed to the other methods. For example, both FixMatch and ReMixMatch use wide ResNet models \cite{huang2017densely} (e.g. Wide ResNet-28-2, Wide ResNet-28-10, etc.) which both perform better than ResNet-50 in their experiments. Our choice of using a ResNet-50 backbone is based on the availability of pretrained version of this network. 
In future work, we will explore other backbone architectures. To address the performance gaps on CIFAR-10, we will also explore methods for generating improved feature representations such as self-supervised approaches that train using pretext tasks on the CIFAR dataset. In addition, we will explore the integration of active learning approaches that selectively request labels for the most informative samples.

\section{Conclusion}
In this paper, our proposed method LiDAM applies deep domain adaptation on artificially induced domain shifts to generate high-quality initial data representations that are used to map pseudo-labels to ground-truth labels. We also introduced a novel iterative matching scheme that progressively improves these predicted data labels and trains a final classifier. To the best of our knowledge, LiDAM is the first approach that utilizes localized domain adaptation to reduce intra-class domain variance for the purpose of boosting self-supervised learning. We experimentally demonstrated the effectiveness of our method on benchmark datasets.

\bibliographystyle{aaai}
\bibliography{main}

\begin{thebibliography}{}

\bibitem[\protect\citeauthoryear{Berthelot \bgroup et al\mbox.\egroup
  }{2019a}]{berthelot2019remixmatch}
Berthelot, D.; Carlini, N.; Cubuk, E.~D.; Kurakin, A.; Sohn, K.; Zhang, H.; and
  Raffel, C.
\newblock 2019a.
\newblock Remixmatch: Semi-supervised learning with distribution alignment and
  augmentation anchoring.
\newblock {\em arXiv preprint arXiv:1911.09785}.

\bibitem[\protect\citeauthoryear{Berthelot \bgroup et al\mbox.\egroup
  }{2019b}]{berthelot2019mixmatch}
Berthelot, D.; Carlini, N.; Goodfellow, I.; Papernot, N.; Oliver, A.; and
  Raffel, C.~A.
\newblock 2019b.
\newblock Mixmatch: A holistic approach to semi-supervised learning.
\newblock In {\em Advances in Neural Information Processing Systems},
  5049--5059.

\bibitem[\protect\citeauthoryear{Cao \bgroup et al\mbox.\egroup
  }{2018a}]{cao2018partial}
Cao, Z.; Long, M.; Wang, J.; and Jordan, M.~I.
\newblock 2018a.
\newblock Partial transfer learning with selective adversarial networks.
\newblock In {\em Proceedings of the IEEE Conference on Computer Vision and
  Pattern Recognition},  2724--2732.

\bibitem[\protect\citeauthoryear{Cao \bgroup et al\mbox.\egroup
  }{2018b}]{cao2018partials}
Cao, Z.; Ma, L.; Long, M.; and Wang, J.
\newblock 2018b.
\newblock Partial adversarial domain adaptation.
\newblock In {\em Proceedings of the European Conference on Computer Vision
  (ECCV)},  135--150.

\bibitem[\protect\citeauthoryear{Caron \bgroup et al\mbox.\egroup
  }{2018}]{caron2018deep}
Caron, M.; Bojanowski, P.; Joulin, A.; and Douze, M.
\newblock 2018.
\newblock Deep clustering for unsupervised learning of visual features.
\newblock In {\em European Conference on Computer Vision}.

\bibitem[\protect\citeauthoryear{Chen \bgroup et al\mbox.\egroup
  }{2020}]{chen2020big}
Chen, T.; Kornblith, S.; Swersky, K.; Norouzi, M.; and Hinton, G.
\newblock 2020.
\newblock Big self-supervised models are strong semi-supervised learners.
\newblock {\em arXiv preprint arXiv:2006.10029}.

\bibitem[\protect\citeauthoryear{Ganin and
  Lempitsky}{2015}]{ganin2015unsupervised}
Ganin, Y., and Lempitsky, V.
\newblock 2015.
\newblock Unsupervised domain adaptation by backpropagation.
\newblock In {\em International conference on machine learning},  1180--1189.

\bibitem[\protect\citeauthoryear{Garg and Kalai}{2017}]{garg2017supervising}
Garg, V.~K., and Kalai, A.
\newblock 2017.
\newblock Supervising unsupervised learning.

\bibitem[\protect\citeauthoryear{He \bgroup et al\mbox.\egroup
  }{2019}]{he2019moco}
He, K.; Fan, H.; Wu, Y.; Xie, S.; and Girshick, R.
\newblock 2019.
\newblock Momentum contrast for unsupervised visual representation learning.
\newblock {\em arXiv preprint arXiv:1911.05722}.

\bibitem[\protect\citeauthoryear{Huang \bgroup et al\mbox.\egroup
  }{2017}]{huang2017densely}
Huang, G.; Liu, Z.; Van Der~Maaten, L.; and Weinberger, K.~Q.
\newblock 2017.
\newblock Densely connected convolutional networks.
\newblock In {\em Proceedings of the IEEE conference on computer vision and
  pattern recognition},  4700--4708.

\bibitem[\protect\citeauthoryear{Krizhevsky, Nair, and
  Hinton}{2009}]{krizhevsky2009cifar}
Krizhevsky, A.; Nair, V.; and Hinton, G.
\newblock 2009.
\newblock Cifar-10 and cifar-100 datasets.
\newblock {\em URl: https://www. cs. toronto. edu/kriz/cifar. html} 6:1.

\bibitem[\protect\citeauthoryear{Laine and Aila}{2016}]{laine2016temporal}
Laine, S., and Aila, T.
\newblock 2016.
\newblock Temporal ensembling for semi-supervised learning.
\newblock {\em arXiv preprint arXiv:1610.02242}.

\bibitem[\protect\citeauthoryear{Lee \bgroup et al\mbox.\egroup
  }{2019}]{Lee_2019_ICCV}
Lee, S.; Kim, D.; Kim, N.; and Jeong, S.-G.
\newblock 2019.
\newblock Drop to adapt: Learning discriminative features for unsupervised
  domain adaptation.
\newblock In {\em The IEEE International Conference on Computer Vision (ICCV)}.

\bibitem[\protect\citeauthoryear{Lee}{2013}]{lee2013pseudo}
Lee, D.-H.
\newblock 2013.
\newblock Pseudo-label: The simple and efficient semi-supervised learning
  method for deep neural networks.
\newblock In {\em Workshop on challenges in representation learning, ICML},
  volume~3.

\bibitem[\protect\citeauthoryear{{Mittal} and {Sangwan}}{2019}]{8776614}
{Mittal}, S., and {Sangwan}, O.~P.
\newblock 2019.
\newblock Big data analytics using machine learning techniques.
\newblock In {\em 2019 9th International Conference on Cloud Computing, Data
  Science Engineering (Confluence)},  203--207.

\bibitem[\protect\citeauthoryear{Miyato \bgroup et al\mbox.\egroup
  }{2018}]{miyato2018virtual}
Miyato, T.; Maeda, S.-i.; Koyama, M.; and Ishii, S.
\newblock 2018.
\newblock Virtual adversarial training: a regularization method for supervised
  and semi-supervised learning.
\newblock {\em IEEE transactions on pattern analysis and machine intelligence}
  41(8):1979--1993.

\bibitem[\protect\citeauthoryear{Sohn \bgroup et al\mbox.\egroup
  }{2020}]{sohn2020fixmatch}
Sohn, K.; Berthelot, D.; Li, C.-L.; Zhang, Z.; Carlini, N.; Cubuk, E.~D.;
  Kurakin, A.; Zhang, H.; and Raffel, C.
\newblock 2020.
\newblock Fixmatch: Simplifying semi-supervised learning with consistency and
  confidence.
\newblock {\em arXiv preprint arXiv:2001.07685}.

\bibitem[\protect\citeauthoryear{Tarvainen and
  Valpola}{2017}]{tarvainen2017mean}
Tarvainen, A., and Valpola, H.
\newblock 2017.
\newblock Mean teachers are better role models: Weight-averaged consistency
  targets improve semi-supervised deep learning results.
\newblock In {\em Advances in neural information processing systems},
  1195--1204.

\bibitem[\protect\citeauthoryear{Wang \bgroup et al\mbox.\egroup
  }{2019}]{wang2019enaet}
Wang, X.; Kihara, D.; Luo, J.; and Qi, G.-J.
\newblock 2019.
\newblock Enaet: Self-trained ensemble autoencoding transformations for
  semi-supervised learning.
\newblock {\em arXiv preprint arXiv:1911.09265}.

\bibitem[\protect\citeauthoryear{Xie \bgroup et al\mbox.\egroup
  }{2019}]{xie2019unsupervised}
Xie, Q.; Dai, Z.; Hovy, E.; Luong, M.-T.; and Le, Q.~V.
\newblock 2019.
\newblock Unsupervised data augmentation for consistency training.
\newblock {\em arXiv preprint arXiv:1904.12848}.

\bibitem[\protect\citeauthoryear{Yu \bgroup et al\mbox.\egroup
  }{2019}]{yu2019transfer}
Yu, C.; Wang, J.; Chen, Y.; and Huang, M.
\newblock 2019.
\newblock Transfer learning with dynamic adversarial adaptation network.
\newblock In {\em 2019 IEEE International Conference on Data Mining (ICDM)},
  778--786.
\newblock IEEE.

\bibitem[\protect\citeauthoryear{Zhang \bgroup et al\mbox.\egroup
  }{2017}]{zhang2017mixup}
Zhang, H.; Cisse, M.; Dauphin, Y.~N.; and Lopez-Paz, D.
\newblock 2017.
\newblock mixup: Beyond empirical risk minimization.
\newblock {\em arXiv preprint arXiv:1710.09412}.

\bibitem[\protect\citeauthoryear{Zhao \bgroup et al\mbox.\egroup
  }{2018}]{zhao2018adversarial}
Zhao, H.; Zhang, S.; Wu, G.; Moura, J.~M.; Costeira, J.~P.; and Gordon, G.~J.
\newblock 2018.
\newblock Adversarial multiple source domain adaptation.
\newblock In {\em Advances in Neural Information Processing Systems},
  8568--8579.

\end{thebibliography}

\end{document}